\pdfoutput=1
\documentclass[10pt,twocolumn,letterpaper]{article}

\usepackage{iccv}
\usepackage{times}
\usepackage{epsfig}
\usepackage{graphicx}
\usepackage{amsmath}
\usepackage{amssymb}

\usepackage{xcolor}
\usepackage{booktabs}
\usepackage{tabularx}
\usepackage{colortbl}
\usepackage{textcomp}
\usepackage{verbatim}
\usepackage{multirow}
\usepackage{makecell}
\usepackage{subcaption}
\usepackage{siunitx}
\usepackage[super]{nth}
\usepackage{authblk}
\usepackage{microtype}

\makeatletter  %% Hack to make PGF/Tikz work with the ICCV style
\@namedef{ver@everyshi.sty}{}
\renewcommand\AB@affilsepx{\hskip 0.15in \protect\Affilfont}

\makeatother
\usepackage{pgf}

\usepackage[pagebackref=true,breaklinks=true,letterpaper=true,colorlinks,bookmarks=false]{hyperref}

\usepackage[capitalize]{cleveref}

 \iccvfinalcopy % *** Uncomment this line for the final submission

\def\iccvPaperID{****} % *** Enter the ICCV Paper ID here

\ificcvfinal\fi

\definecolor{LightGray}{gray}{0.95}

\usepackage{color}
\usepackage[absolute]{textpos}
\definecolor{gray92}{gray}{0.92}

\begin{document}

%%%%%%%%% TITLE
\title{A Strong Baseline for the VIPriors Data-Efficient Image Classification Challenge}

\author[1]{Bj{\"o}rn Barz\thanks{{\tt\footnotesize bjoern.barz@uni-jena.de}}}
\author[2]{Lorenzo Brigato\thanks{{\tt\footnotesize brigato@diag.uniroma1.it}}}
\author[2]{Luca Iocchi}
\author[1]{Joachim Denzler}
\affil[1]{Friedrich Schiller University Jena}
\affil[2]{Sapienza University of Rome}

\maketitle
\ificcvfinal\thispagestyle{empty}\fi

%%%%%%%%% ABSTRACT
\begin{abstract}
Learning from limited amounts of data is the hallmark of intelligence, requiring strong generalization and abstraction skills.
In a machine learning context, data-efficient methods are of high practical importance since data collection and annotation are prohibitively expensive in many domains.
Thus, coordinated efforts to foster progress in this area emerged recently, e.g., in the form of dedicated workshops and competitions.
Besides a common benchmark, measuring progress requires strong baselines.
We present such a strong baseline for data-efficient image classification on the VIPriors challenge dataset, which is a sub-sampled version of ImageNet-1k with 100 images per class.
We do not use any methods tailored to data-efficient classification but only standard models and techniques as well as common competition tricks and thorough hyper-parameter tuning.
Our baseline achieves 69.7\% accuracy on the VIPriors image classification dataset and outperforms 50\% of submissions to the VIPriors 2021 challenge.
\end{abstract}

%%%%%%%%% BODY TEXT
\section{Introduction}
\label{sec:intro}

Many recent advances in computer vision and machine learning in general have been achieved by large-scale pre-training on massive datasets \cite{dosovitskiy2021vit,cui2018large}.
Methodical improvements of the recent past such as transformer architectures \cite{vaswani2017attention} followed this data-driven paradigm by reducing inductive priors and hence providing more degrees of freedom to the model.
However, many applications do not have access to millions of samples due to the high cost of data collection and annotation, e.g., in the domain of medical images.
In such a scenario, deep learning methods need to incorporate inductive priors drawing from human prior knowledge to enable better generalization from less data.

The research community in this area recently started to organize workshops and competitions on visual inductive priors for data-efficient deep learning (VIPriors) \cite{bruintjes2021vipriors} to encourage such methodical advances and compare the competing methods on a common benchmark dataset.
In the case of the VIPriors challenge, this benchmark is a subset of the ImageNet-1k dataset \cite{russakovsky2015imagenet} with 100 images per class.

While such a challenge eliminates the effect of varying training datasets on the comparison of methods, many other factors remain that can overshadow methodical progress.
For example, simple data augmentation can have a high impact on the classification performance in a data-deficient setting \cite{brigato2021close}.
Furthermore, a recent study found that many published works neglect hyper-parameter tuning for the baselines against which they compare, unfairly biasing the evaluation \cite{brigato2021deic}.
Especially in the context of competitions, additional tricks such as ensembling are often applied to improve the final results without methodical contributions.
Therefore, the availability of a strong baseline to compare against is crucial for assessing how far the field of data-efficient image classification has come compared to standard techniques.

We provide such a strong baseline using a deep neural network, state-of-the-art data augmentation \cite{cubuk2020randaugment}, thorough hyper-parameter optimization on held-out validation data, and usual competition tricks such as test-time augmentation \cite{shanmugam2020tta} and ensembling.
We submitted our model to the image classification track of the VIPriors 2021 competition, which is the second iteration of the VIPriors challenge.
In this competitive environment, our baseline (69.7\%) surpassed the baseline provided by the organizers (31.2\%) substantially, outperformed 50\% of the submissions, and falls only 5 percent points behind the winning submission.

In the following, we first briefly re-introduce the VIPriors challenge dataset in \cref{sec:dataset} and then provide all technical details of our training and inference pipeline in \cref{sec:methods}.
The results and an ablation study analyzing the contribution of individual parts of the pipeline are presented in \cref{sec:results}.

%------------------------------------------------------------------------
\section{Dataset}
\label{sec:dataset}

We train our model on the VIPriors Image Classification Challenge dataset\footnote{\url{https://github.com/VIPriors/vipriors-challenges-toolkit/tree/master/image-classification}}, which is a subset of ImageNet-1k \cite{russakovsky2015imagenet}.
It comprises images from 1,000 classes of everyday objects and natural scenes collected from the web using image search engines.
While the original dataset contains over a million images with different numbers of samples for each class, the VIPriors challenge selects exactly 100 images per class.
Due to the large number of 1,000 classes, the reduced dataset still contains 100,000 images in total, which is a rather large number compared to other data-efficient image classification benchmarks \cite{brigato2021deic}.

The dataset is split into 50,000 training and 50,000 validation images.
We use this split for hyper-parameter optimization and train the final models on all 100,000 images.
The original test set of ImageNet-1k comprising another 50,000 images is used for evaluating the models submitted to the VIPriors challenge.

%-------------------------------------------------------------------------
\section{Classification Pipeline}
\label{sec:methods}

In the following, we detail the individual steps of our training and inference pipeline.
None of these is particularly tailored towards data-efficient deep learning.
The resulting performance can hence be considered a baseline for the VIPriors challenge.

\subsection{Data Pre-Processing and Augmentation} 
We normalize all images by subtracting the channel-wise mean and dividing by the standard deviation computed on the combined training and validation data.

For state-of-the-art data augmentation, we apply Randaugment \cite{cubuk2020randaugment}, which can be adapted to the data at hand by controlling two hyper-parameters $N$ and $M$.
Randaugment picks exactly $N$ consecutive transformations for each image at random from a pool of 16 transformations such as rotation, contrast change, color shift, shear etc.
The parameter $M$ controls the maximum strength of the distortion.

After distorting the image using Randaugment, we apply scale augmentation using the \texttt{RandomResizedCrop} transform from PyTorch\footnote{\url{https://pytorch.org/vision/stable/transforms.html\#torchvision.transforms.RandomResizedCrop}} as follows:
A crop with a random aspect ratio drawn from $[\frac{3}{4}, \frac{4}{3}]$ and an area between 10\% and 100\% of the original image area is extracted from the image and then resized to $224 \times 224$ pixels.

%%%%%%%%% Table HPO
\begin{table}[t]
    \renewcommand{\arraystretch}{1.3}
    \setlength{\tabcolsep}{5pt}
    \begin{tabularx}{\linewidth}{Xcc}
        \toprule
        Hyper-Parameter &      Search Space     &     Value     \\
        \midrule
        Learning Rate   & loguniform(1e-4, 0.1) & \num{1.98e-3} \\
        \rowcolor{LightGray}
        Weight Decay    & loguniform(1e-5, 0.1) & \num{4.21e-4} \\
        $N$             &      \{1, 2, 3\}      & 2 \\
        \rowcolor{LightGray}
        $M$             &    \{2, 6, 10, 14\}   & 14 \\
        Batch Size      &       \{8, 16\}       & 8 \\
        \bottomrule
    \end{tabularx}
    \caption{Hyper-parameter search space and values.}
    \label{tab:hpo_params}
\end{table}

\subsection{Architecture and Optimization}
Since deep residual networks \cite{he2016deep} are still a popular choice for image classification models, we use the ResNeXt-101 architecture \cite{xie2017resnext}.
One could argue that such a deep network is over-parameterized for a data-deficient setting.
In preliminary experiments with a shallower ResNet-50 \cite{he2016deep}, however, we observed a substantial advantage of the deeper ResNeXt-101 (see \cref{subsec:ablation}).

For training the model, we use the standard categorical cross-entropy loss and stochastic gradient descent (SGD) with a momentum of 0.9 and weight decay.
The learning rate follows a cosine annealing schedule \cite{loshchilov2016sgdr}, which reduces it smoothly during the training process.
The initial learning rate and the weight decay factor are tunable hyper-parameters.

\subsection{Hyper-Parameter Optimization}

We tune the hyper-parameters of our pipeline on the training and validation sets of the challenge dataset, which are disjoint from the test set used for final performance evaluation (see \cref{sec:dataset}).
Values to be tested for the initial learning rate and weight decay are sampled from a log-uniform space.
The batch size as well as the hyper-parameters $N$ and $M$ of Randaugment are chosen from pre-defined sets.
Details about the search space and the final values determined during hyper-parameter optimization are provided in \cref{tab:hpo_params}.

For selecting hyper-parameter configurations to be tested and scheduling experiments, we employ Asynchronous HyperBand with Successive Halving (ASHA) \cite{li2020system} as implemented in the Ray library\footnote{\url{https://docs.ray.io/en/master/tune/}}.
This search algorithm exploits parallelism and aggressive early-stopping to tackle large-scale hyper-parameter optimization problems.
Trials are evaluated and stopped based on their accuracy on the validation split.

Two main parameters need to be configured for the ASHA algorithm: the number of trials and the grace period.
The former controls the number of hyper-parameter configurations tried in total while the latter the minimum time after which a trial can be stopped.
We use a total of 200 trials and a grace period of 10 epochs.
The maximum number of epochs was set to 200.

\subsection{Final Training}
After having determined suitable hyper-parameters using the procedure described above, we train the network with the determined configuration on the combined training and validation split.
We furthermore double the number of training epochs to 400 for the final training compared to the hyper-parameter optimization step.

\begin{figure}[t]
    \centering
    \resizebox{\linewidth}{!}{\input{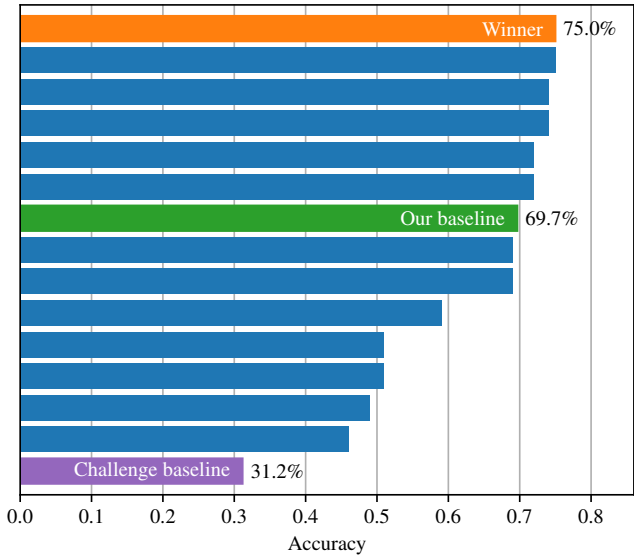}}
    \caption{Results of the VIPriors 2021 Image Classification Challenge.}
    \label{fig:results}
\end{figure}

\subsection{Test-Time Augmentation}

A common trick to improve the performance of an already trained image classifier is test-time augmentation (TTA) \cite{shanmugam2020tta}.
The idea is to derive multiple variants from a single image, make a prediction for each of them, and aggregate the results into a single class prediction.

In our case, we derive $3 \times 5 \times 2 = 30$ variants from each test image.
First, we create 3 resized versions so that the smaller side of the image is 256, 288, and 352 pixels large, respectively.
From each of these 3 differently scaled images, we then extract 5 crops of size $224 \times 224$, one from the center and one from each corner.
Furthermore, a horizontally flipped version of each crop is added to the set of derived images.

After obtaining the logits predicted by the model for each of the 30 variants, we apply the softmax activation to each prediction vector and take the average over all image variants to obtain the final prediction vector for the original image.

\subsection{Ensemble}

Another popular trick often employed in competitions for boosting prediction performance consists in combining multiple independently trained models to an ensemble and aggregating their predictions.
To this end, we train 20 instances of our model with different random initializations and select the class with the highest average class score (after TTA) as the finally predicted one.

%------------------------------------------------------------------------
\section{Results}
\label{sec:results}

\subsection{Challenge Results}
\label{subsec:challenge-results}

Our image classification pipeline achieves a top-1 accuracy of 69.66\% on the test set of the VIPriors Image Classification Challenge.
As can be seen in \cref{fig:results}, this performance falls only 5 percent points behind the winning submission, even though we only use standard techniques.

The accuracy obtained by a single model without ensembling and inference-time tricks such as TTA is 61.15\% on average.
This is almost twice as high as the baseline provided by the challenge organizers, which was trained with untuned hyper-parameters.

\subsection{Ablation Study}
\label{subsec:ablation}

\begin{figure}[t]
    \centering
    \resizebox{\linewidth}{!}{\input{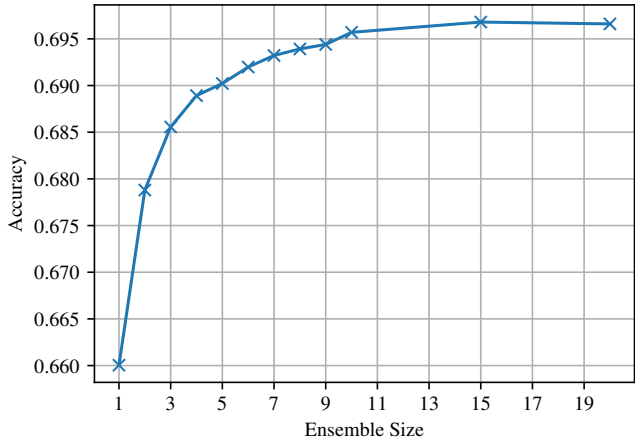}}
    \caption{Effect of ensemble size on classification accuracy.}
    \label{fig:ensemble-size}
\end{figure}

\begin{figure*}[t]
    \centering
    \resizebox{.96\linewidth}{!}{\input{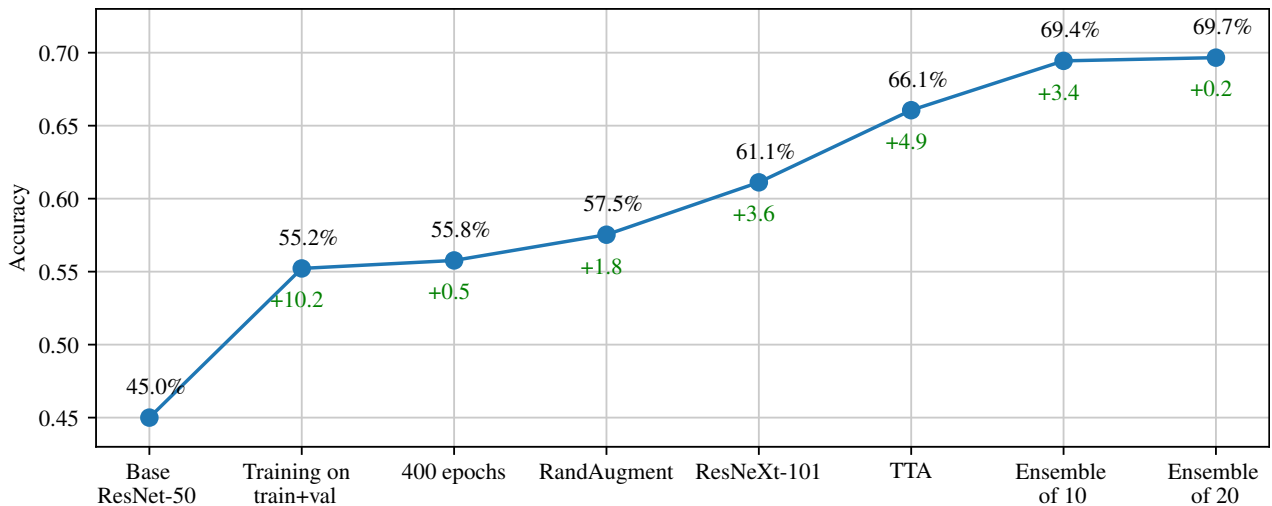}}
    \caption{Ablation study: Effect of each incremental addition to the classification pipeline.}
    \label{fig:ablation}
\end{figure*}

We analyze the impact of the individual modifications and additions made to our classification pipeline in \cref{fig:ablation}.

Unsurprisingly, doubling the amount of training data by conducting the final training on the combined training and validation set increases the performance the most by 10 percent points.
Doubling the training time, in contrast, only results in a minor improvement by half a percent point.

Recent studies indicate that lower-complexity models provide better generalization performance in data-deficient scenarios than more complex models \cite{brigato2021close}.
This does not seem to be the case for the VIPriors dataset, which is still rather large.
We initially experimented with a ResNet-50 architecture but then switched to a deeper ResNeXt-101, which improved the performance by 3.6 percent points.

TTA had the second-largest impact after training on more data with an improvement of 5 percent points.
Ensembling contributed another 3.6 percent points.
However, an ensemble of 20 models performs only 0.2 percent points better than 10 models.
The effect of ensemble size is depicted in more detail in \cref{fig:ensemble-size}.
The gains diminish clearly when the ensemble grows beyond 8-10 models.

%------------------------------------------------------------------------
\section{Conclusions}
\label{sec:conclusions}

We have provided a strong baseline for the VIPriors Image Classification Challenge.
Using only standard techniques, we outperformed 50\% of submissions in the 2021 edition of the challenge.
The most useful ingredients of our classification pipeline were strong data augmentation using Randaugment \cite{cubuk2020randaugment}, a deep ResNeXt-101 architecture \cite{xie2017resnext}, test-time augmentation \cite{shanmugam2020tta}, and ensembling.

This baseline allows for obtaining a realistic picture of the progress in the field of data-efficient image classification.
The top three methods all outperform our baseline by four to five percent points, which is significant on the ImageNet-1k test set but still not too far from the baseline.

{\small
\bibliographystyle{ieee_fullname}
\bibliography{egbib}
}

\end{document}